\documentstyle[epsf]{kbsjrnl} 
\newcommand{\VCdim}{{\mbox{VCdim}}}
\newcommand{\h}{{\mathbold{h}}}
\newcommand{\Pb}{{\mathbold{P}}}
\newcommand{\scPb}{{\mathbold{\scriptstyle P}}}
\newcommand{\er}{{\mbox{er}}}
\newcommand{\erh}{{\hat{\er}}}
\newcommand{\doteqdot}{\doteq}

\newcommand{\bb}{}
\newcommand{\eqref}[1]{(\ref{#1})}

\newcommand{\mathbold}[1]{\mbox{\boldmath $\bf#1$}}

\renewcommand{\glossary}[2]{}
\def\addcontentsline#1#2#3{} 
\newcommand{\thickapprox}{\doteqdot}
\newtheorem{thm}{Theorem}

\newtheorem{defn}{Definition}

\newcommand{\ie}{{\em i.e.\ }}

\newcommand{\eg}{{\em e.g.\ }}

\renewcommand{\max}{{\text{max}}}

\newcommand{\ptt}{p(z|\thetat)}

\newcommand{\E}{\bb E}
\newcommand{\HH}{{\mathbold{H}}}

\renewcommand{\H}{{\cal H}}

\renewcommand{\P}{{\cal P}}

\newcommand{\Zn}{{Z^n}}

\newcommand{\znm}{{z^{(n,m)}}}
\newcommand{\znmm}{{z^{(n,m-1)}}}
\newcommand{\z}{{\mathbold{z}}}

\newcommand{\be}{\begin{equation*}}

\newcommand{\ep}{{\varepsilon}}

\newcommand{\C}{{\cal C}}
\newcommand{\N}{{\bb N}}

\renewcommand{\(}{\left(}
\renewcommand{\)}{\right)}
\renewcommand{\[}{\left[}
\renewcommand{\]}{\right]}
\newcommand{\text}[1]{\mbox{\rm #1}}

\newcommand{\Cbar}{{\overline C}}
\newcommand{\Lbar}{{\overline L}}

\renewcommand{\hbar}{{\overline h}}

\newcommand{\thetat}{{\tilde{\theta}}}

\newcommand{\R}{{\bb R}}

\begin{document}
\begin{article}

     
\authorrunninghead{}
\titlerunninghead{}
     
\setcounter{page}{71} 

\title{Theoretical Models of Learning to Learn\thanks{This work was
	supported in part by EPSRC grants \#K70366 and \#K70373}}
     
\author{Jonathan Baxter}
\email{jon@syseng.anu.edu.au} 
     
\affil{Department of Systems Engineering \\ Research School of
Information Science and Engineering \\ Australian National University \\
Canberra 0200 \\ Australia}
     
     
\abstract{
A Machine can only learn if it is biased in some way. Typically the
bias is supplied by hand, for example through the choice of an appropriate 
set of features. However, if the learning machine is embedded within an {\em
environment} of related tasks, then it can {\em learn} its own bias by
learning sufficiently many tasks from the environment
\cite{95a,97b}. In this paper two models of bias learning (or
equivalently, learning to learn) are introduced and the main
theoretical results presented. The first model is a PAC-type model
based on empirical process theory, while the second is a hierarchical
Bayes model.}

\section{Introduction}
Hume's analysis \cite{Hume1737} shows that there is no {\em a priori} basis for
induction. In a machine learning context, this means that a learner
must be biased in some way for it to generalise well \cite{mitchell91}.
Typically such bias is introduced by hand through the skill and
insights of experts, but despite many notable successes, this process
is limited by the experts' abilities. Hence a desirable goal
is to find ways of automatically {\em learning} the bias. 
Bias learning is a form of {\em learning to learn}, and the two
expressions will be used interchangeably throughout this document.

The purpose of this chapter is to present an overview of two models of
{\em supervised} bias learning. The first \cite{95a,95c} is based
on Empirical Process theory (henceforth the {\em EP} model) and the
second \cite{97b} is based on Bayesian inference and information theory
(henceforth the {\em Bayes} model). Empirical process
theory is a general theory that includes the analysis of pattern
classification first introduced by Vapnik and Chervonenkis
\cite{vapnik71a,vapnik82a}. Note that these are models of {\em
supervised} bias learning and as such have little to say about
learning to learn in a reinforcement learning setting.

In this introduction a high level overview of the features common to
both models will be presented, and then in later sections the details
and main results of each model will be discussed.

In ordinary models of machine learning the learner is presented with a
{\em single} task. Learning the ``right bias'' in such a model does
not really make sense, because the ultimate bias is one which
completely solves the task. Thus in single-task learning, bias
learning or learning to learn is the same as learning.

In order to learn bias one has introduce extra assumptions about the
learning process. The central assumption of both the Bayes model and
the EP model of bias learning is that the learner is embedded within an {\em
environment} of related problems. The learner's task
is to find a bias that is appropriate for the entire environment, not
just for a single task.

A simple example of an environment of learning problems with a common bias is
handwritten character recognition. A preprocessing stage that
identifies and removes any (small) rotations, dilations and
translations of an image of a character will be advantageous for
recognising all characters.  If the set of all individual character
recognition problems is viewed as an environment of learning tasks,
this preprocessor represents a bias that is appropriate to all tasks
in the environment. 

Preprocessing can also be viewed as feature extraction, and there are
many classes of learning problems that possess common feature
sets. For example, one can view face recognition as a collection of
related learning problems, one for each possible face classifier, and
it is likely that there exists sets of features that are good for
learning all faces. A similar conclusion applies to other domains such
as speech recognition (all the individual word classifiers may be
viewed as separate learning problems possessing a common feature set),
fingerprint recognition, and so on. The classical approach to
statistical pattern recognition in these domains is to first {\em
guess} a set of features and then to learn each problem by estimating
a simple (say linear) function of the features. The choice of features
represents the learner's bias, thus in bias learning the goal is to
get the learner to {\em learn} the features instead of guessing them.

In order to perform a theoretical analysis of bias learning, we assume
the tasks in the environment are generated according to some
underlying probability distribution. For example, if the learner is
operating in an environment where it must learn to recognise faces,
the distribution over learning tasks will have its support restricted
to face recognition type problems. The learner acquires information
about the environment by sampling from this distribution to generate
multiple learning problems, and then sampling from each learning
problem to generate multiple training sets.  The learner can then
search for bias that is appropriate for learning all the tasks.

In the EP model, the learner is provided with a family  of
hypothesis spaces and it searches for an hypothesis space that
contains good solutions to all the training sets. Such a hypothesis
space can then be used to learn {\em novel} tasks drawn from the same
environment. The key result of the EP model (theorem \ref{EPthm} in
section \ref{EPsec}) gives a bound on the number of tasks and number
of examples of each task required to ensure that a hypothesis space
containing good solutions to all training sets will, with high
probability, contain good solutions to novel tasks drawn from the same
environment. This ability to learn novel tasks after seeing
sufficiently many examples of sufficiently many tasks is the formal
definition of learning to learn under the EP model.

The Bayes model is the same as the EP model in that the learner is
assumed to be embedded within an environment of related tasks and can
sample from the environment to generate multiple training sets
corresponding to different tasks. However, the Bayes bias learner
differs in the way it uses the information from the multiple training
sets.  In the Bayes model, the distribution over learning tasks in the
environment is interpreted as an {\em objective} prior
distribution. The learner does not know this distribution, but does
have some idea of a set $\Pi$ of possible prior distributions to which the
true distribution belongs. The learner starts out with a {\em
hyper-prior} distribution on $\Pi$ and based on the data in the
training sets, updates the hyper-prior to a {\em hyper-posterior}
using Bayes' rule. The hyper-posterior is then used as a prior
distribution when learning novel tasks. In section \ref{bayes} results
will be presented showing how the {\em information} needed to learn
each task (in a Shannon sense) decays to the minimum possible for the
environment as the number of tasks and number of examples of each
tasks seen already grows. Within the Bayes model, this is the formal
definition of learning to learn. 

Before moving on to the details of these models, it is worth pausing
to assess what bias learning solves, and what it doesn't---and in a
sense can {\em never}---solve.
On face value, being able to learn the right bias appears to violate
Hume's conclusion that there can be no {\em a priori} basis for
induction. However this is not the case, for the bias learner learner
is still fundamentally limited by the possible choices of bias
available. For example, if a learner is learning a set of features for
an environment in which there are in fact no small feature sets, then any
bias it comes up with (\ie any feature set) will be a very poor bias
for that environment. Thus, there is still guesswork involved in
determining the appropriate way to {\em hyper-bias} the learner. The
main advantage of bias learning is that this hyper-bias can be much
weaker than the bias: the right hyper-bias for many environments is just
that there exists a set of features, whereas specifying the right bias
means actually finding the features. 

\section{Statistical Models of Ordinary Learning}

To understand how bias learning can be modeled from a statistical
perspective, it is necessary to first understand how ordinary learning
is modeled from a statistical perspective. The empirical process (EP)
process approach and the Bayes approach will be discussed in turn.

\subsection{The empirical process (EP) approach}
\label{epord}
The empirical process (EP) approach to modeling ordinary
(single-task) learning has the following essential ingredients:
\begin{itemize}
\item An {\em input space} $X$ and an {\em output space} $Y$,
\item a {\em probability distribution} $P$ on $X\times Y$,
\item a {\em loss function} $l\colon Y\times Y\to \R$, and
\item a {\em hypothesis space} $\H$ which is a set of {\em hypotheses}
or functions $h\colon X\to Y$.
\end{itemize}
As an example, if the problem is to learn to recognize images of
Mary's face using a neural network, 
then $X$ would be the set of all images (typically
represented as a subset of $\R^d$ where each component is a pixel
intensity), $Y$ would be the set $\{0,1\}$, and the distribution $P$
would be peaked over images of different faces and the correct class
labels. The learner's hypothesis space $\H$ would be a class of neural
networks mapping the input space ($\R^d$) to $\{0,1\}$.
\begin{equation}
l(y,y') := \left\{ \begin{array}{ll}
			1& \mbox{if $y\neq y'$} \\
			0& \mbox{if $y = y'$}
			\end{array} \right.	
\end{equation}
Using the loss function allows us to present a unified treatment of both
concept learning ($Y = \{0,1\}$, $l$ as above), and real-valued
function learning (\eg regression) in which $Y=\R$ and $l(y,y') =
(y-y')^2$. 

The goal of the learner is to select a hypothesis $h\in \H$ with
minimum {\em expected loss}: 
\begin{equation}
\label{expordloss}
\er_P(h) := \int_{X\times Y} l(h(x),y)\, dP(x,y).
\end{equation}
For classifying Mary's face, the $h\in \H$ with minimum value of
$\er_P(h)$ is the one that makes the fewest number of mistakes on average.
Of course, the learner does not know $P$ and so it cannot search
through $\H$ for an $h$ minimizing $\er_P(h)$. In practice, the
learner samples repeatedly from the distribution $P$ to generate a
{\em training set} 
\begin{equation}
z:= \{(x_1,y_1),\dots,(x_m,y_m)\},
\end{equation}
and instead of minimizing $\er_P(h)$, the learner searches for an
$h\in \H$ minimizing the {\em empirical loss} on sample $z$:
\begin{equation}
\label{empordloss}
\erh_z(h) := \frac1m\sum_{i=1}^m l(h(x_i),y_i).
\end{equation}
Of course, there are more intelligent things to do with the data than
simply minimizing empirical error---for example one can add
regularisation terms to avoid over-fitting. However we do not consider
those issues here as they do not substantially alter the discussion.

Minimizing $\erh_z(h)$ is only a sensible thing to do if there is some
guarantee that $\erh_z(h)$ is close to expected loss $\er_P(h)$.  This
will in turn depend on the ``richness'' of the class $\H$ and the size
of the training set ($m$). If $\H$ contains every possible function
then clearly there can never be any guarantee that $\erh_z(h)$ is
close to $\er_P(h)$. 
The conditions ensuring convergence between
$\erh_z(h)$ and  $\er_P(h)$ are by now well understood; in the case
Boolean function learning ($Y= \{0,1\}$), convergence is controlled by
$\VCdim(\H)$---the {\em VC-dimension} of $\H$ (see \eg
\cite{AB,vapnik82a}). 
The following is typical of the theorems in this
area.
\begin{thm}
\label{vcthm}
Let $P$ be any probability distribution on $X\times \{0,1\}$ and
suppose $z= \{(x_1,y_1),\dots,(x_m,y_m)\}$ is generated by sampling
$m$ times from $X\times \{0,1\}$ according to $P$. Let $d :=
\VCdim(\H)$.  Then with probability at least $1-\delta$ (over the
choice of the training set $z$), {\bf all} $h\in \H$ will satisfy
\begin{equation}
\label{vceq}
\er_P(h) \leq \erh_z(h) + \[\frac{32}m\(d\ln\frac{2 e m}{d} +
\ln\frac4\delta\)\]^{1/2}
\end{equation}
\end{thm}
There are a number of key points about this theorem:
\begin{enumerate} 
\item We can never say for certain that $\er_P(h)$ and $\erh_z(h)$ are
close, only that they are close with high probability
($1-\delta$). This is because no matter how large the training set,
there is always a chance that we will get unlucky and generate a
highly unrepresentative sample $z$. 
\item Keeping the confidence parameter $\delta$ fixed, and ignoring
$\log$ factors, \eqref{vceq} shows that the difference between the
empirical estimate $\erh_z(h)$ and the true loss $\er_P(h)$ decays like
$\sqrt{d/m}$, {\em uniformly} for all $h\in \H$. Thus, for
sufficiently large training sets $z$ and if $d = \VCdim(\H)$ is finite, we
can be confident that an $h$ with small empirical error will
generalise well.
\item \label{note} If $\H$ contains an $h$ with zero error, and the
learner always chooses an $h$ consistent with the training set, then
the rate of convergence of $\erh_z(h)$ and $\er_P(h)$ can be improved
to $d/m$.
\item Often results such as theorem \ref{vcthm} are called {\em
uniform convergence} results, because they provide bounds for all
$h\in \H$.  
\end{enumerate}
Theorem \ref{vcthm} only provides conditions under which the
deviation between $\er_P(h)$ and $\erh_z$  is small, it does not
guarantee that the true error $\er_P(h)$ will actually be small. This
is governed by the choice of $\H$. If $\H$ contains a solution with
small error and the learner minimizes error on the training set, then
with high probability $\er_P(h)$ will be small. However, a bad choice
of $\H$ will mean there is no hope of achieving small error. Thus, the
{\em bias} of the learner in the EP model is represented by the choice
of $\H$. 

\subsection{The Bayes approach}
\label{bayesord}
The Bayes approach and the EP approach are not all that different. In
fact, the EP approach can be understood as a maximum likelihood
approximation to the Bayes solution.
The essential ingredients of the Bayes approach
to modeling ordinary (single-task) learning are:
\begin{itemize}
\item An {\em input space} $X$ and an {\em output space} $Y$,
\item a {\em set} of probability distributions $P_\theta$ on $X\times
Y$, parameterised by $\theta\in \Theta$, and
\item a {\em prior} distribution $p(\theta)$ on $\Theta$.
\end{itemize}
The hypothesis space $\H$ of the EP model has been replaced with a set
of distributions $\{P_\theta\colon \theta\in\Theta\}$. 
As with the EP model, the learning task is represented by a
distribution $P_{\theta^*}$ on $X\times Y$, only this time we assume
{\em realizability}, \ie that $\theta^*\in\Theta$. The prior
distribution $p(\theta)$ represents the learner's initial beliefs
about the relative plausibility of each $P_\theta$. The richness of
the set $\{P_\theta\colon\theta\in \Theta\}$ and the prior $p(\theta)$
represent the bias of the learner.

\sloppy
Again the learner does not know $\theta^*$, but has access to a
training set $z=\{(x_1,y_1),\dots,(x_m,y_m)\}$ of pairs $(x_i,y_i)$
sampled according to $P_{\theta^*}$. Upon receipt of the data $z$, the
learner updates its prior distribution $p(\theta)$ to a {\em posterior}
distribution $p(\theta|z)$ using Bayes rule:
\begin{eqnarray}
\label{bayesrule}
p(\theta|z) &:=& \frac{p(z|\theta) p(\theta)}{p(z)} \\
&=& \frac{\prod_{i=1}^m p(x_i,y_i|\theta) p(\theta)}{\int_\Theta
p(z|\theta) p(\theta) \, d\theta}. \nonumber
\end{eqnarray}
Often we are not interested in modeling the input distribution, only
the conditional distribution on $Y$ given $X$. In that case,
$p(x,y|\theta)$  will factor into $p(x) p(y|x;\theta)$. The
posterior distribution $p(\theta|z)$ can be used to predict the output
$y$ of a novel input $x^*$ by averaging:
\begin{equation}
\label{average}
p(y|x^*;z) := \int_\Theta p(y|x^*;\theta) p(\theta|z)\, d\theta.
\end{equation}

One would hope that as the data increases, predictions made in this
way would become increasingly accurate. There are many ways to measure
what we mean by ``accurate'' in this setting. The one considered here
is the {\em Kullback-Liebler} (KL) divergence between the true distribution
$P_{\theta^*}$ on $X\times Y$, and the posterior distribution $P_m$ on
$X\times Y$ with density
\begin{equation}
p(x,y|z) :=  \int_\Theta p(x,y|\theta) p(\theta|z)\, d\theta.
\end{equation}
The KL divergence between $P_{\theta^*}$ and $P_m$ is defined to be
\begin{equation}
\label{KL}
D_K(P_\theta^*\|P_m) := \int_{X\times Y} p(x,y|\theta^*) \log 
\frac{p(x,y|\theta^*)}{p(x,y|z)}\, dx\,dy
\end{equation}
Note that if $p(x,y|\theta) = p(x)p(y|\theta)$ as above then 
\begin{equation}
D_K(P_\theta^*\|P_m) = \int_{X\times Y} p(x) p(y|x;\theta^*) \log 
\frac{p(y|x; \theta^*)}{p(y|x;z)}\, dx\,dy.
\end{equation}
This form of the KL divergence has a natural interpretation: it is
(within one bit) the expected extra number of bits needed to encode
the output $y$ using a code generated by the posterior $p(y|x;z)$,
over and above what would be required using an optimal code (one
generated from $p(y|x;\theta^*)$). The expectation is over all pairs
$(x,y)$ drawn according to the true distribution $P_{\theta^*}$. This
quantity is only zero if the posterior is equal to the true
distribution.

In \cite{CB1,CB2} an analysis of $D_K(P_{\theta^*}\|P_m)$ was given for
the limit of large training set size $(m)$. They showed that if
$\Theta$ is a compact subset of $\R^d$, and under certain extra
restrictions which we won't discuss here:
\begin{equation}
\label{CB}
D_K(P_{\theta^*}\|P_m) = \frac{d}{m} + o\(\frac1m\) 
\end{equation}
where $o(1/m)$ stands for a function $f(m)$ for which $mf(m)
\rightarrow 0$ as $m\rightarrow\infty$.

There is a strong similarity between this result and theorem
\ref{vcthm} in the zero error case (see note \ref{note} after the
theorem). $D_K(P_{\theta^*}\|P_m)$ is the analogue of $|\er_z(h) -
\er_P(h)|$ in this case, but because we have assumed realizability,
$\er_z(h) = 0$. So theorem \ref{vcthm} says that choosing any
hypothesis consistent with the data will guarantee you an error of no
more than $d/m$, where $d$ is the VC dimension of the learner's
hypothesis space. Although the error measure is different, equation 
\eqref{CB} says essentially the same thing: if you classify novel data
using a posterior generated according to Bayes rule, you will suffer
an error of no more than $d/m$, where now $d$ is the dimension of the
$\Theta$. 

\section{The empirical process (EP) model of learning to learn}
\label{EPsec}
Recall from the introduction that the main extra assumption of both
the Bayes and EP models of bias learning is that the learner is
embedded in an {\em environment} of related tasks, and can sample from
the environment to generate multiple training sets belonging to
multiple different tasks. In the EP model of ordinary (single-task)
learning, the learning problem is represented by a distribution $P$ on
$X\times Y$. So in the EP model of bias learning, an environment of
learning problems is represented by a pair $(\P,Q)$ where $\P$ is the
set of all probability distributions on $X\times Y$ (\ie $\P$ is the
set of all possible learning problems), and $Q$ is a distribution on
$\P$\footnote{Strictly speaking, in order for $Q$ to be well defined
we need to specify a $\sigma$-algebra of subsets of $\P$. However,
such considerations are beyond the scope of the present
discussion. See \cite{95c} for more details.}. 
$Q$ controls which
learning problems the learner is likely to see. For example, if the
learner is in a face recognition environment, $Q$ will be highly
peaked over face-recognition-type problems, whereas if the learner is
in a character recognition environment $Q$ will be peaked over
character-recognition-type problems.

Recall from the end of section \ref{epord} that the learner's bias is
represented by its choice of hypothesis space $\H$. So to enable the
learner to learn the bias, it is supplied with a {\em family} or set of
hypothesis spaces $\HH := \{\H\}$. As each $\H$ is itself a set of
functions $h\colon X\to Y$, $\HH$ is a {\em set of sets of functions}.

Putting this together, formally a {\em learning to learn} or {\em bias
learning} problem consists of:
\begin{itemize} 
\item an {\em input space} $X$ and an {\em output space} $Y$,
\item a {\em loss function} $l\colon Y\times Y\to \R$,
\item an {\em environment} $(\P,Q)$ where $\P$ is the set of all
probability distributions on $X\times Y$ and $Q$ is a distribution on
$\P$,
\item a {\em hypothesis space family} $\HH = \{\H\}$ where each $\H\in\HH$
is a set of functions $h\colon X\to Y$. 
\end{itemize}
The goal of a bias learner is to find a hypothesis space $\H\in\HH$
minimizing the loss (recall equation \eqref{expordloss})
\begin{eqnarray}
\label{exploss}
\er_Q(\H) &:=&\int_\P\inf_{h\in \H} \er_P(h)\,dQ(P) \\
&=&\int_\P\inf_{h\in
\H} \int_{X\times Y} l(h(x), y)\, dP(x,y) \,dQ(P). \nonumber
\end{eqnarray}
The only way $\er_Q(\H)$ can be small is if, with high
$Q$-probability, $\H$ contains a good solution to any problem $P$
drawn at random according to $Q$. In this sense $\er_Q(\H)$ measures
how appropriate the bias embodied by $\H$ is for the environment
$(\P,Q)$. 

In general the learner will not know $Q$, so it will not be able to
find an $\H$ minimizing $\er_Q(\H)$ directly. However, the learner can
sample from the environment in the following way:
\begin{itemize}
\item Sample $n$ times from $\P$ according to $Q$ to yield:\\
$P_1,\dots,P_n$.
\item Sample $m$ times from $X\times Y$ according to each $P_i$ to yield:\\
$z_i = \{(x_{i1},y_{i1})\dots,(x_{im},y_{im})\}$.
\item The learner receives the {\em $(n,m)$-sample}:
\begin{equation}
\begin{array}{ccccc}
& (x_{11},y_{11}) & \cdots & (x_{1m}, y_{1m}) & = z_1 \\
\z:= &  \vdots & \ddots & \vdots & \vdots\\ 
& (x_{n1},y_{n1}) & \cdots & (x_{nm}, y_{nm}) & = z_n
\end{array}
\end{equation}
\end{itemize}
Note that an $(n,m)$-sample is simply $n$ training sets
$z_1,\dots,z_n$ sampled from $n$ different learning tasks
$P_1,\dots,P_n$, where each task is selected according to the
environmental probability distribution $Q$.

Instead of minimizing $\er_Q(\H)$, the learner searches for an
$\H\in\HH$ minimizing the {\em empirical loss} on the sample $\z$,
where this is defined by:
\begin{eqnarray}
\label{emploss}
\erh_\z(\H) &:=&\frac1n\sum_{i=1}^n
\inf_{h\in \H}\er_{z_i}(h) \\
&=&\frac1n\sum_{i=1}^n
\inf_{h\in \H}\frac1m\sum_{j=1}^m l\(h_i(x_{ij}), y_{ij}\) \nonumber
\end{eqnarray}
(recall equation \eqref{empordloss}). Note that $\erh_\z(\H)$ is a
biased estimate of $\er_Q(\H)$. 

The question of generalisation within this framework now becomes:
How many tasks ($n$) and how many examples of each task ($m$) do
we need to ensure that $\erh_\z(\H)$ and $\er_Q(\H)$ are close with high
probability? 

Or, informally, how many tasks and how many examples of each task are
required to ensure that a hypothesis space with good solutions to all
the training tasks will contain good solutions to novel tasks drawn
from the same environment?

In order to present the main theorem answering this question, some
extra definitions must be introduced.
\begin{defn}
For any hypothesis $h\colon X\to Y$, define 
$h_l\colon X\times Y\to \R$ by 
\begin{equation}
h_l(x,y) := l(h(x),y)
\end{equation}
For any hypothesis space $\H$ in the hypothesis space family $\HH$, 
define 
\begin{equation}
\H_l:= \{h_l\colon h\in \H\}.
\end{equation}
For any sequence of $n$ hypotheses $(h_1,\dots,h_n)$, define 
$(h_1,\dots,h_n)_l\colon (X\times Y)^n\to \R$ by 
\begin{equation}
(h_1,\dots,h_n)_l(x_1,y_1,\dots,x_n,y_n) := \frac1n\sum_{i=1}^n l(h_i(x_i),y_i).
\end{equation}
We will also use $\h_l$ to denote $(h_1,\dots,h_n)_l$.
For any $\H$ in the hypothesis space family $\HH$, define 
\begin{equation}
\H^n_l  := \{(h_1,\dots,h_n)_l\colon h_1,\dots,h_n\in \H\}.
\end{equation}
Define
\begin{equation} 
\HH^n_l := \bigcup_{\H\in\HH} \H^n_l.
\end{equation}
\end{defn}
In the first part of the definition above, hypotheses $h\colon X\to Y$
are turned into functions $h_l$ mapping $X\times Y\to \R$  using the loss
function. $\H_l$ is then just the collection of all such
functions where the original hypotheses come from $\H$. $\H_l$ is
often called a {\em loss-function class}. In our case we are
interested in the average loss across $n$ tasks, where each of the
$n$ hypotheses is chosen from a fixed hypothesis space $\H$. This
motivates the definition of $\h_l$ and $\H^n_l$. Finally,
$\HH^n_l$ is the collection of all $(h_1,\dots,h_n)_l$, with the
restriction that all $h_1,\dots,h_n$ belong to a single hypothesis
space $\H\in\HH$.
\begin{defn}
For each $\H\in \HH$, define $\H^*\colon \P\to \R$ by 
\begin{equation}
\H^*(P) := \inf_{h\in \H} \er_P(h).
\end{equation}
For the hypothesis space family $\HH$, define 
\begin{equation} 
\HH^* := \{\H^*\colon \H\in \HH\}.
\end{equation}
\end{defn} 
It is the ``size'' of $\HH^n_l$ and $\HH^*$ that controls how large 
the $(n,m)$-sample $\z$ must be to ensure $\er_\z(\H)$ and $\er_Q(\H)$ are
close uniformly over all $\H\in\HH$. Their size will be defined in
terms of certain covering numbers, and in order to define the covering
numbers, we need first to define how to measure the distance between
elements of $\HH^n_l$ and also between elements of $\HH^*$.
\begin{defn}
Let $\Pb = (P_1,\dots,P_n)$ be any sequence of $n$ probability
distributions on $X\times Y$.
For any $\h_l,\h'_l\in \HH^n_l$, define
\begin{splitmath}
\label{dp}
d_\scPb(\h_l,\h'_l) := \int_{(X\times Y)^n}
|\h_l(x_1,y_1,\dots,x_n,y_n) &-
\h'_l(x_1,y_1,\dots,x_n,y_n)| \\
\,
dP_1(x_1,y_1) \dots dP_n(x_n,y_n)
\end{splitmath}
\noindent For any $\H_1^*,\H_2^*\in \HH^*$, define
\begin{equation}
\label{dq}
d_Q(\H_1^*,\H_2^*) := \int_\P \left|\H_1^*(P) - \H_2^*(P)\right | \,dQ(P)
\end{equation}
\end{defn}
It is easily verified that $d_\scPb$ is a {\em pseudo-metric} on
$\HH^n_l$, and similarly that $d_Q$ is a pseudo-metric on $\HH^*$. A
pseudo-metric is simply a metric without the condition that $\rho(x,y)
= 0 \Rightarrow x = y$. For example, $d_Q(\H^*_1,\H^*_2)$ could equal
$0$ simply because the distribution $Q$ puts mass one on some
distribution $P$ for which $\H^*_1(P) = \H^*_2(P)$, and not because
$\H_1^* = \H_2^*$.

\begin{defn}
\label{covdef}
An $\ep$-cover of $(\HH^*,d_Q)$ is a set $\{\H^*_1,\dots,\H^*_N\}$ such
that for all $\H^*\in\HH^*$, $d_Q(\H^*,\H_i) \leq \ep$ for some $i=1\dots
N$. Let $\N(\ep,\HH^*,d_Q)$ denote the size of the smallest such cover.
Set 
\begin{equation}
\label{bub}
\C(\ep,\HH^*) := \sup_{Q} \N(\ep,\HH^*,d_Q).
\end{equation}
We can define $\N(\ep,\HH^n_l,d_\scPb)$ in a similar way, using
$d_\scPb$ in place of $d_Q$. Again, set:
\begin{equation}
\label{bib}
\C(\ep,\HH^n_l) := \sup_{\scPb} \N(\ep,\HH^n_l,d_\scPb).
\end{equation}
\end{defn}
\noindent Now we have enough machinery to state the main theorem.
\begin{thm}
\label{EPthm}
Let $Q$ be any probability distribution on $\P$, the set of all
distributions on $X\times Y$. Suppose $\z$ is an $(n,m)$-sample 
generated by sampling $n$ times from $\P$ according to $Q$ to give
$P_1,\dots,P_n$, and then sampling $m$ times from each $P_i$ to
generate $z_i = \{(x_{i1},y_{i1}),\dots,(x_{im},y_{im})\}$, $i=1,\dots,n$.
Suppose the loss function $l\colon Y\times Y\to \R$ has range $[0,1]$
(any bounded loss function can be rescaled so this is true).
Let $\HH=\{\H\}$ be any hypothesis space family. 
If the number of tasks $n$ satisfies 
\begin{equation}
\label{nbound}
n \geq \frac{288}{\ep^2}\ln\frac{8\C\(\frac{\ep}{48},\HH^*\)}\delta,
\end{equation}
and the number of examples $m$ of each task satisfies
\begin{equation}
\label{mbound}
m \geq
\max\left\{\frac{288}{n\ep^2}\ln\frac{8\C(\frac{\ep}{48},\HH^n_l)}\delta,
\frac{18}{\ep^2}\right\}
\end{equation}
then with probability at least $1-\delta$ (over the $(n,m)$-sample
$\z$), all $\H\in\HH$ will satisfy
\begin{equation}
\label{blip}
\er_Q(\H) \leq \erh_\z(\H) + \ep
\end{equation}
\end{thm}
\noindent For a proof of a similar theorem to this one, see the proof
of theorem 7 in \cite{95c}. Note that the constants in this theorem
have not been heavily optimized. \\

\noindent There are several important points to note about theorem
\ref{EPthm}:
\begin{enumerate}
\item In order to learn to learn (in the sense that $\er_Q(\H)$ and 
$\erh_\z(\H)$ are close uniformly over all $\H\in\HH$), both the
number of tasks $n$ and the number of examples of each task $m$ must
be sufficiently large.
\item We can never say for certain that $\er_Q(\H)$ and $\erh_\z(\H)$ are
close, only that they are close with high probability
($1-\delta$). Regardless of the size of the $(n,m)$
sample $\z$, we still might get unlucky and generate unrepresentative
learning problems $P_1,\dots,P_n$ or unrepresentative examples of
those learning problems, although the chance of being unlucky
diminishes as $m$ and $n$ grow. 
\item Once the learner has found an $\H\in\HH$ with a small value of
$\erh_\z(\H)$, it will then use $\H$ to learn novel tasks $P$ drawn
according to $Q$. Assuming that the learner is always able to find an
$h\in \H$ minimizing $\er_P(h)$, theorem \ref{EPthm} tells us that
with probability at least $1-\delta$, the expected value of $\er_P(h)$
on a novel task $P$ will be less than $\erh_\z(\H) + \ep$. Of course,
this does not rule out really bad performance on some tasks
$P$. However, the probability of generating such ``bad'' tasks can be
bounded. In particular, note that $\er_Q(\H)$ is just the expected
value of the function $\H^*$ over $\P$, and so by Markov's inequality,
for $\gamma > 0$, 
\begin{eqnarray*}
\Pr\left\{P\colon \inf_{h\in\H} \er_P(h) \geq \gamma\right\} &=&
\Pr\left\{P\colon \H^*(P) \geq \gamma\right\} \\ 
&\leq& \frac{\E_Q \H^*}{\gamma} \\ 
&=&\frac{\er_Q(\H)}{\gamma} \\ 
&\leq& \frac{\erh_\z(\H) + \ep}{\gamma}\quad \text{(with probability
$1-\delta$)} 
\end{eqnarray*}
\item Keeping the accuracy and confidence parameters $\ep,\delta$
fixed, note that the number of examples of each task required for good
generalisation obeys
\begin{equation}
m = O\(\frac1n\ln \C\(\ep,\HH^n_l\)\).
\end{equation}
So provided $\ln \C\(\ep,\HH^n_l\)$ increases sublinearly with $n$, the
upper bound on the number of examples required of each task will {\em
decrease} as the number of tasks increases. This is discussed further
after theorem \ref{ntaskthm} below.
\end{enumerate}

Theorem \ref{EPthm} only provides conditions under which 
$\erh_\z(\H)$ and $\er_Q(\H)$ are close, it does not guarantee
that $\er_Q(\H)$ is actually small. This is governed by the choice of
$\HH$. If $\HH$ contains a hypothesis space $\H$ with a small value of
$\er_Q(\H)$, and the learner minimizes error on the $(n,m)$ sample
$\z$, then with high probability $\er_Q(\H)$ will be small. However,
a bad choice of $\HH$ will mean there is no hope of finding an $\H$
with small error. In this sense the choice of $\HH$ represents the
{\em hyper-bias} of the learner. 

It may seem that we have simply replaced the problem of selecting the
right bias (\ie selecting the right hypothesis space $\H$) with the
equally difficult problem of selecting the right hyper-bias (\ie the
right hypothesis space family $\HH$). However, in many cases selecting
the right hyper-bias is far easier than selecting the right bias. For
example, in section \ref{featsec} we will see how the feature selection
problem may be viewed as a bias selection problem. Selecting the right
features can be extremely difficult if one knows little about the
environment, but specifying only that a set of features
should exist and then learning those features is far simpler. 

\subsection{Learning multiple tasks}
\label{nsec}
It may be that the learner is not interested in learning to learn, but
simply wants to learn $n$ tasks from the environment $(\P,Q)$.  As in
the previous section, we assume the learner starts out with a
hypothesis space family $\HH$, and also that it receives an
$(n,m)$-sample $\z$ generated from the $n$ distributions
$P_1,\dots,P_n$. This time, however, the learner is simply looking for
$n$ hypotheses $(h_1,\dots,h_n)$, all contained in the same hypothesis
space $\H$, such that the average training set error of the $n$ hypotheses is
minimal. Denoting $(h_1,\dots,h_n)$ by $\h$, this error is defined by 
\begin{eqnarray}
\label{empnerr}
\erh_\z(\h) &:=&\frac1n\sum_{i=1}^n\erh_{z_i}(h_i) \\
             &=&\frac1n\sum_{i=1}^n\frac1m\sum_{j=1}^m l(h_i(x_{ij}),
             y_{ij})\nonumber
\end{eqnarray} 
For any hypothesis space $\H$, let $\H^n:= \{(h_1,\dots,h_n)\colon
h_i\in\H, i=1,\dots,n\}$. Let $\HH^n := \cup_{\H\in \HH} \H^n$. $\HH^n$ is
simply the set of all possible sequences $(h_1,\dots,h_n)$ where all
the $h_i's$ come from the same hypothesis space $\H$ (recall the
definition of $\HH^n_l$ for a similar concept).
Writing $\Pb = (P_1,\dots,P_n)$, the learner's generalisation error
in this context is measured by the average generalisation error across
the $n$ tasks:
\begin{eqnarray}
\label{expnerr}
\er_\scPb(\h) &:=&\frac1n\sum_{i=1}^n \er_{P_i}(h_i) \\
	     &=& \frac1n\sum_{i=1}^n \int_{X\times Y} l(h_i(x),y)\,
	     dP_i(x,y) \nonumber
\end{eqnarray}
Recall definition \ref{covdef} for the meaning of $\C(\ep,\HH^n_l)$.
\begin{thm}
\label{ntaskthm}
Let $\Pb = (P_1,\dots,P_n)$ be $n$ probability distributions on
$X\times Y$ and let $\z$ be an $(n,m)$-sample generated by sampling
$m$ times from $X\times Y$ according to each $P_i$. 
Suppose the loss function $l\colon Y\times Y\to \R$ has range $[0,1]$
(any bounded loss function can be rescaled so this is true).
Let $\HH=\{\H\}$ be any hypothesis space family. 
If the number of examples $m$ of each task satisfies
\begin{equation}
\label{nmbound}
m \geq
\max\left\{\frac{72}{n\ep^2}\ln\frac{4\C(\frac{\ep}{24},\HH^n_l)}\delta,
\frac{18}{\ep^2}\right\}
\end{equation}
then with probability at least $1-\delta$ (over the choice of 
$\z$), any $\h\in\HH^n$ will satisfy
\begin{equation}
\label{blipp}
\er_\scPb(\h) \leq \erh_\z(\h) + \ep
\end{equation}
\end{thm}
\noindent Notes:
\begin{enumerate}
\item Note that the bound on $m$ in theorem \ref{ntaskthm} is
virtually identical to the bound on $m$ in theorem \ref{EPthm}. 
\item The important thing about the bound on $m$ is that it depends
{\em inversely} on the number of tasks $n$ (assuming that the first
part of the ``max'' expression is the dominate one). 
In fact, it is easy to
show that for any hypothesis space family $\HH$, 
\begin{equation}
\C\(\ep,\HH^1_l\) \leq \C\(\ep,\HH^n_l\) \leq \C\(\ep,\HH^1_l\)^n.
\end{equation}
Thus 
\begin{equation}
\label{fart}
\ln \C\(\ep,\HH^1_l\) \leq \ln \C\(\ep,\HH^n_l\) \leq n\ln
\C\(\ep,\HH^1_l\).
\end{equation}
So keeping the accuracy parameters $\ep$ and $\delta$ fixed, and
plugging \eqref{fart} into \eqref{nmbound}, we see that the upper
bound on the number of examples required of each task never {\em
increases} with the number of tasks, and at best decreases as
$O\(\frac1n\)$. Although only an upper bound, this provides a strong
hint that learning multiple related tasks should be advantageuos on a
``number of examples required per task'' basis.
\item In section \ref{featsec} it will be shown that all types of
behaviour, from no advantage at all to $O(\frac1n)$ decrease, are possible.
\end{enumerate}

\subsection{Feature learning with neural networks}
\label{featsec}
Consider the following quote:
\begin{quote}
The classical approach to estimating multidimensional functional
dependencies is based on the following belief:

Real-life problems are such that there exists a small number of
``strong features,'' simple functions of which (say linear
combinations) approximate well the unknown function. Therefore, it
is necessary to carefully choose a low-dimensional feature space and
then to use regular statistical techniques to construct an
approximation.
\end{quote}
(from ``The Nature of Statistical Learning Theory'', Vapnik 1996.) 
It must be pointed out that Vapnik advocates an alternative approach
in his book: that of using an extremely large number of simple
features but choosing a hypothesis with maximum classification
margin. However his approach cannot be viewed as a form of bias
learning or learning to learn, whereas the strong feature approach
can, so here we will concentrate on the latter.

The aim of this section is to use the ideas of the previous section to
show how neural-network feature sets can be {\em learnt} for an
environment of related tasks.  

\begin{figure}
\begin{center}
\leavevmode
\epsfysize=3in\epsfbox{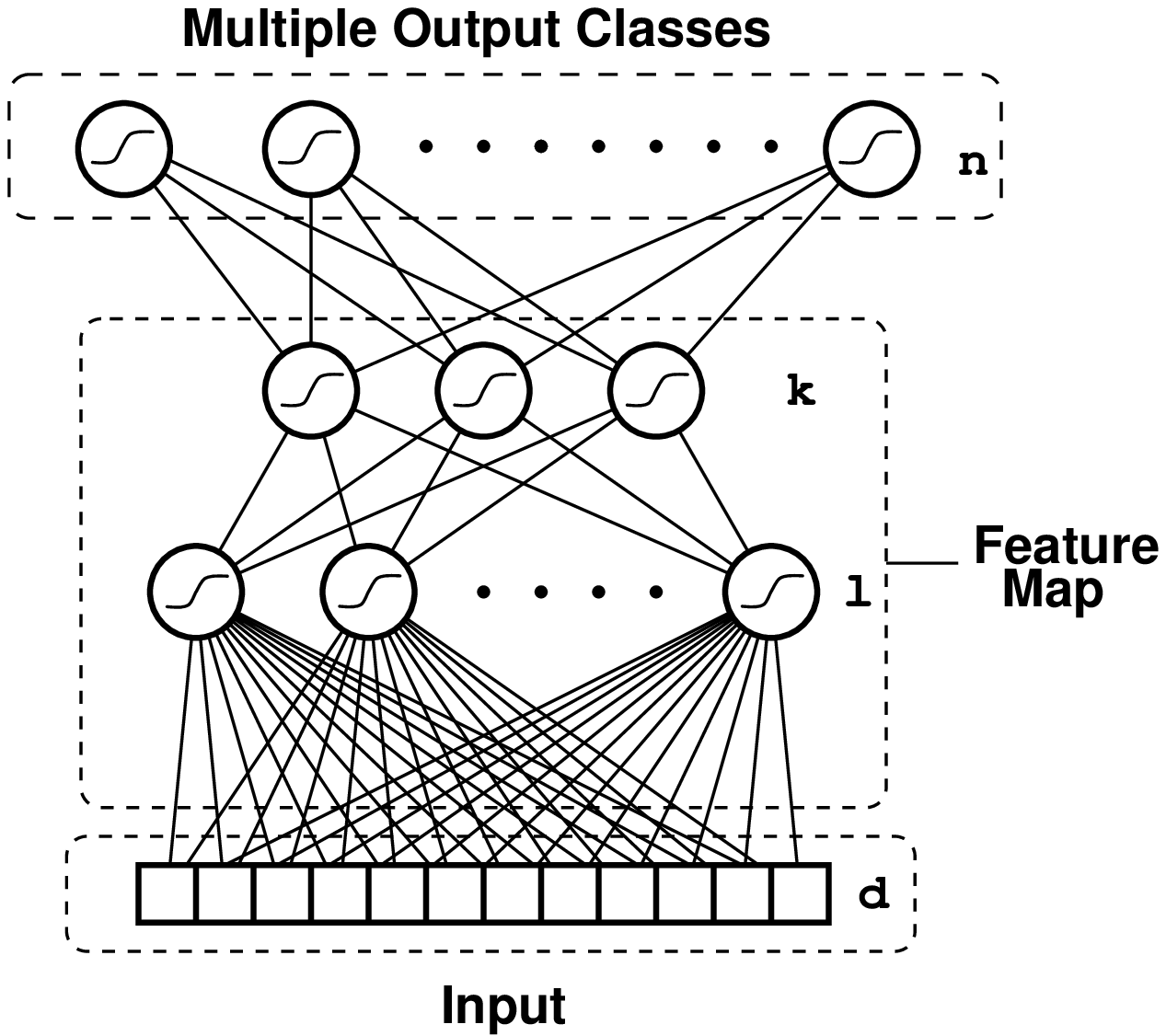}
\caption{\label{nnet} Neural network for feature learning. The feature
map is implemented by the first two hidden layers. The $n$ output
nodes correspond to the $n$ different tasks in the $(n,m)$-sample
$\z$.}
\end{center}
\end{figure}

In general, a set of features may be viewed as a map from the
(typically high-dimensional) input space $\R^d$ to a much smaller
dimensional space $\R^k$ ($k\ll d$). Any such bounded feature map can
be approximated to arbitrarily high accuracy by a one-hidden-layer
neural network with $k$ output nodes. This is illustrated in Figure
\ref{nnet}. Fixing the number of nodes in the first hidden layer, let
$\Phi_w\colon \R^d\to \R^k$ denote the feature map computed by the the
neural network with weights $w$. The set of all such feature maps is
$\{\Phi_w\colon w\in \R^W\}$ where $W$ is the number of weights in
the first two layers.

For argument's sake, assume the ``simple functions'' of the features
(mentioned in the above quote) are squashed linear maps. Denoting the
$k$ components of the feature map $\Phi_w$ by
$\phi_{w,1},\dots,\phi_{w,k}$, each setting of the feature map weights
generates a hypothesis space $\H_w$ by
\begin{equation}
\label{hysp}
\H_w:= \left\{\sigma\(\sum_{i=1}^k \alpha_i \phi_{w,i}\)\colon
(\alpha_1,\dots,\alpha_k) \in \R^k\right\},
\end{equation}
where $\sigma\colon\R\to \R$ is the squashing function. $\H_w$ is
simply the set of all squashed linear functions of the features
$\Phi_w$. 
The set of all such hypothesis spaces,
\begin{equation}
\HH := \{\H_w\colon w\in \R^W\} 
\end{equation}
is a hypothesis space family. 

Finding the right set of features for the environment $(\P,Q)$ is
equivalent to finding the right hypothesis space $\H_w\in \HH$. 

As in the previous section, the correct set of features may be learnt
by finding a hypothesis space with small error on a sufficiently large
$(n,m)$-sample $\z$ (recall that an $(n,m)$-sample is simply $n$
training sets corresponding to $n$ different learning
tasks). Specializing to squared loss, in the present framework the
error of $\H_w$ on $\z$ (equation \eqref{emploss}) is given by
\begin{equation}
\label{featemploss}
\erh_\z(\H_w) = \frac1n\sum_{i=1}^n \inf_{(\alpha_1,\dots,\alpha_k)\in
\R^k}
\frac1m\sum_{j=1}^m \[\sigma\(\sum_{l=1}^k \alpha_l
\phi_{w,l}(x_{ij})\) - y_{ij}\]^2 
\end{equation}
Using gradient descent and an $n$ output node network as in figure
\ref{nnet}, output weights $(\alpha_1,\dots,\alpha_k)$ and feature
weights $w$ minimizing \eqref{featemploss} can be found. For details
see \cite{95a}. 

The size of $\z$ ensuring that the resulting features will be good for
learning novel tasks from the same environment is given by theorem
\ref{EPthm}. All we have to do is compute the logarithm of the
covering numbers $\C(\ep,\HH^n_l)$ and $\C(\ep,\HH^*)$.  If the
feature weights $w$ and the output weights $\alpha_1,\dots,\alpha_k$
are bounded, and the squashing function $\sigma$ is
Lipschitz\footnote{$\sigma$ is Lipschitz if there exists a constant
$K$ such that $\sigma(x,x') \leq K|x-x'|$ for all $x,x'\in\R$.}, then
there exist constants $\kappa,\kappa'$ (independent of $\ep,W$ and
$k$) such that for all $\ep > 0$,
\begin{eqnarray*}
\ln\C(\ep,\HH^n_l) &\leq& 2\(kn + W\)\ln\frac{\kappa}{\ep} \\
\ln\C(\ep,\HH^*) &\leq& 2 W\ln\frac{\kappa'}{\ep}
\end{eqnarray*}
(see \cite{thesis} for a proof).
Plugging these expressions into theorem \ref{EPthm} gives the
following theorem.
\begin{thm}
\label{featthm}
Let $\HH = \{\H_w\}$ be a hypothesis space family where each hypothesis
space $\H_w$ is a set of squashed linear maps composed with a neural
network feature map, as above. Suppose the number of features is
$k$, and the total number of feature weights is W. Assume
all feature weights and output weights are bounded, and the squashing
function $\sigma$ is Lipschitz. Let $\z$ be an $(n,m)$-sample
generated from the environment $(\P,Q)$.  If
\begin{equation}
n \geq O\(\frac1{\ep^2}\[W + \log\frac1\delta\]\),
\end{equation}
and
\begin{equation}
m \geq O\(\frac1{\ep^2}\[k + \frac1n\(W + \log\frac1\delta\)\]\)
\end{equation}
then with probability at least $1 - \delta$ any 
$\H_w$ will satisfy
\begin{equation}
\er_Q(\H_w) \leq \er_\z(\H_w) + \ep.
\end{equation}
\end{thm}
\noindent Notes:
\begin{enumerate}
\item Keeping the accuracy paramters $\ep$ and $\delta$ fixed, the
upper bound on the number of examples required of each task behaves
like $O(k + W/n)$. The same upper bound also applies in theorem
\ref{ntaskthm}.
\item Once the feature map is learnt, only the output weights have to
be estimated to learn a novel task. Again keeping the accuracy
parameters fixed, this requires no more that $O(k)$ examples. 
\item Thus, as the number of tasks learnt increseas, the upper bound
on the number of examples required of each task decays to the minimum
possible, $O(k)$.
\item If the ``small number of strong features'' assumption is
correct, then $k$ will be small. However, typically we will have very
little idea of what the features are, so the size of the feature
network will have to be huge, so $W\gg k$. 
\item $O(k + W/n)$ decreases most rapidly with increasing $n$ when
$W\gg k$, so at least in terms of the upper bound on the number of
examples required per task, learning small feature sets is an ideal
application for learning to learn.
\item Note that if we do away with the feature map altogether then
$W=0$ and the upper bound on $m$ becomes $O(k)$, independent of $n$.
So in terms of the upper bound, learning $n$ tasks becomes just as
hard as learning one task. At the other extreme, if we fix the output
weights then effctively $k=0$ and the number of examples required of
each task decreases as $O(W/n)$. Thus a range of  behaviour in the number of
examples required of each task is possible: from no improvement at all
to an $O(1/n)$ decrease as the number of tasks $n$ increases.
\item To rigorously conclude that learning $n$ tasks is better than
learning one, we would have to show a matching {\em lower bound} of
$\Omega(k + W/n)$ on the number of examples required of each
task. Rather than search for lower bounds within the EP model (which
are difficult to prove), we discuss a Bayes model of learning to
learn in the next section where simultaneous upper and lower bounds
appear more naturally. 
\end{enumerate}

\section{The Bayes model of learning to learn}
\label{bayes}

Recall from section \ref{bayesord} that in Bayesian models of ordinary
learning the learner's bias is represented by the space of possible
distributions $\{P_\theta\colon \theta\in\Theta\}$ along with the
choice of prior $p(\theta)$. The learning task $P$ is assumed to be
equal to some $P_{\theta^*}$ where $\theta^*\in \Theta$. 

Observe that $p(\theta)$ is a {\em subjective} prior distribution over
a set of distributions $\{P_\theta\}$. It is subjective because it
simply reflects the prior beliefs of the learner, not some objective
stochastic phenomenon. Now note that the environment $(\P,Q)$ consists
of a set of distributions $\P$, and a distribution $Q$ on $\P$, and
furthermore that $\P$ can be sampled according to $Q$ to generate
multiple tasks $P_1,\dots,P_n$ (recall the discussion in section
\ref{EPsec}).  This makes $Q$ an {\em objective} prior distribution.
Objective in the sense that it can be sampled, \ie it corresponds to
some objective stochastic phenomenon.

If we assume $\P = \{P_\theta\colon \theta\in \Theta\}$ (so now $\P$
is a restricted subset of all possible distributions on $X\times Y$),
the goal of a bias learner in this framework is to find the right prior
distribution $Q$. To do this, the learner must have available a set of
possible prior distributions $\{P_\pi\colon\pi\in\Pi\}$ from which to
choose. Each $P_\pi$ is a distribution on $\Theta$. We assume
realizability, so that $Q = P_{\pi^*}$ for some $\pi^*\in\Pi$. 

To summarize, the Bayes model of learning to learn consists of the
following ingredients:
\begin{itemize}
\item An {\em input space} $X$ and an {\em output space} $Y$,
\item a set of probability distributions $P_\theta$ on $X\times Y$,
parameterised by $\theta\in\Theta$,
\item a set of {\em prior} distributions $P_\pi$ on $\Theta$,
parameterised by $\pi\in\Pi$.
\item an {\em objective} or {\em environmental} prior distribution
$P_{\pi^*}$ where $\pi^*\in\Pi$.
\item To complete the model, the learner also has a subjective {\em
hyper-prior} distribution $P_\Pi$ on $\Pi$.
\end{itemize} 

The two-tiered structure with a set of possible priors $\{P_\pi\colon
\pi\in\Pi\}$ and a hyper-prior $p(\pi)$ on $\Pi$ makes this model an
example of a {\em hierarchical Bayesian model}
\cite{berger86,Good80}. 

As with the EP model, the learner receives an $(n,m)$-sample $\z$,
generated by first sampling $n$ times from $\Theta$ according
$P_{\pi^*}$ to give $\theta_1,\dots,\theta_n$, and then sampling $m$
times from each $X\times Y$ according to each $P_{\theta_i}$ to
generate $z_i = (x_{i1},y_{i1}),\dots,(x_{im},y_{im})$. To simplify
the notation in this section, let $Z:= X\times Y$ and $z_{ij} :=
(x_{ij},y_{ij})$. As it will be necessary to distinguish between
$(n,m-1)$ samples and $(n,m)$ samples, this will be made explicit in
the notation by writing $\znm$ instead of $\z$:
\begin{equation}
\znm =
\begin{array}{ccc}
z_{11} & \cdots & z_{1m} \\ \vdots & \ddots & \vdots \\ z_{n1} &
\cdots & z_{nm}
\end{array}
\end{equation}

\subsection{Loss as the extra information 
required to predict the next observation}
\label{lossec}
Recall that in the Bayes model of single task learning (section
\ref{bayesord}), the learner's loss was measured by the amount of
extra information needed to encode novel examples of the task. So one way
to measure the advantage in learning $n$ tasks together is by the rate
at which the learner's loss in predicting novel examples decays for
each task. This question is similar to that addressed by theorem
\ref{ntaskthm}. 

So fix the number of tasks $n$, sample $n$ tasks $\theta^n =
\theta_1,\dots,\theta_n$ according to the true prior
$P_{\pi^*}$, and then for each $m=1,2,\dots$ the learner has
already seen $m-1$ examples of each task:
\begin{equation}
\znmm =
\begin{array}{ccc}
z_{11} & \cdots & z_{1m-1}\\ \vdots & \ddots &
\vdots \\ 
z_{n1} &\cdots & z_{nm-1}
\end{array}
\end{equation}
where each row is drawn according to $P^{m-1}_{\theta_i}$ (or
equivalently, each column is drawn according to $P_{\theta^n}$).
The learner then:
\begin{itemize}
\item generates the posterior distribution $p(\theta^n|\znmm)$ 
on the set of all $n$
tasks, $\Theta^n$, according to Bayes' rule:
\begin{eqnarray}
p(\theta^n|\znmm) &=& \frac{p(\znmm|\theta^n) p(\theta^n)}{p(\znmm)} \nonumber	\\ &=&
\frac{p(\theta^n) \prod_{i=1}^n \prod_{j=1}^{m-1}
p(z_{ij}|\theta_i)}{p(\znmm)}
\label{blurb}
\end{eqnarray}
where $p(\znmm) = \int_{\Theta^n} p(\theta^n) \prod_{i=1}^n \prod_{j=1}^{m-1}
p(z_{ij}|\theta_i) \, d\theta^n$.
\item uses the posterior distribution to generate a predictive
distribution on $Z^n$,
\begin{equation}
p(z^n|\znmm) = \int_{\Theta^n} p(z^n|\theta^n) p(\theta^n|\znmm) \,
d\theta^n,
\end{equation}
\item and suffers a loss, $\Lbar_{n,m}$, equal to the expected amount
of extra information needed {\em per task} to encode a novel example
of each task using the predictive distribution $p(z^n|\znmm)$, over
and above the minimum amount of information, \ie the information 
required using the true distribution, $p(z^n|\theta^n)$:
\begin{equation}
\label{hellop}
\Lbar_{n,m} := \frac1n\E_{\Zn|\theta^n}
\log\frac{p(z^n|\theta^n)}{p(z^n|\znmm)}.
\end{equation}
\end{itemize}
Note that the loss at the first trial is:
\begin{equation}
\Lbar_{n,1} := \E_{\Zn|\theta^n} \log\frac{p(z^n|\theta^n)}{p(z^n)},
\end{equation}
where $p(z^n)$ is the learner's initial distribution on $\Zn$ before
any data has arrived,
\begin{equation}
p(z^n) = \int_{\Theta^n} p(z^n|\theta^n) p(\theta^n)\,d\theta^n =
\int_\Pi\int_{\Theta^n} p(z^n|\theta^n)
p(\theta^n|\pi)\,d\theta^n\, p(\pi)\,d\pi
\end{equation}
To understand better the meaning of $\Lbar_{n,m}$, consider
the loss associated with learning a single classification task.  In
this case $Z=X\times\{0,1\}$. If we assume that only the conditional
distribution on class labels is affected by the model, then
$p(z|\theta) = p(x)p(y|x,\theta)$, and for the predictive
distribution, $p(z|z^m) = p(x)p(y|x,z^m)$.  Let $\alpha(x) :=
p(y=1|x,\theta)$ and $\beta(x) := p(y=1|x,z^m)$. Substituting these
expressions into \eqref{hellop} and simplifying yields
\begin{equation}
\Lbar_{1,m} = \E_{X} \[\alpha(x) \log\frac{\alpha(x)}{\beta(x)}
+ (1-\alpha(x)) \log\frac{1-\alpha(x)}{1-\beta(x)}\].
\end{equation}
The expression in square brackets is zero if $\alpha(x) = \beta(x)$,
\ie if the conditional distributions on class labels are the same for
the true and predictive distributions. It increases slowly as
$\alpha(x)$ and $\beta(x)$ diverge.
  
It turns out that $\Lbar_{n,m}$ is difficult to analyse, so instead 
we look at the cumulative loss over a sequence of trials:
\begin{equation}
\label{num}
\Cbar_{n,m,\pi^*} := \sum_{k=0}^{m-1} \E_{\Theta^n|\pi^*}
\E_{Z^{(n,k)}|\theta^n} \Lbar_{n,k+1},
\end{equation}
\ie the {\em expected} total loss incurred by the learner after $m$
steps of the above process. Note that the expectation is over all
sequences of $n$ tasks $\theta^n$ drawn according to $P_{\pi^*}$ and
all $(n,k)$-samples drawn according to $p(z^n|\theta^n)$.

\subsection{\mathbold{(a,b)} models}
In \cite{97b}, the asymptotic behaviour of $\Cbar_{n,m,\pi^*}$ as a
function of $m$ was analysed for general hierarchical models. To
illustrate the results, and to show how they apply to the feature
learning example of section \ref{featsec}, here we will restrict our
attention to $(a,b)$-models. The concept of an $(a,b)$-model was
formally defined in \cite{97b} as follows:
\begin{defn}
\label{abdef}
An $(a,b)$-model is a hierarchical model in which $\Pi=\R^b$,
$\Theta=\R^a\times \R^b$ and the following conditions hold:
\begin{enumerate}
\item The priors $p(\theta|\pi)$ are of the form
\begin{equation}
\label{exam}
p(\theta = (x^a,x^b) | \pi) = \delta(x^b - \pi) g_\pi(x_a)
\end{equation}
where $\delta(\cdot)$ is the $b$-dimensional Dirac delta function and
$g_\pi$ is a continuous function on $\R^a$ that also varies
continuously with $\pi$. 
\item The hyper-prior on $\Pi$, $P_\Pi$ has continuous density $p(\pi)$ and
the true prior $\pi^*$ has positive density $p(\pi^*)$.
\item The conditional distributions $p(z|\theta)$ are
twice continuously differentiable functions of $\theta$.
\end{enumerate}
\end{defn}
The definition of an $(a,b)$-model in \cite{97b} contains two
technical conditions which have been omitted in the above
definition. The interested reader is referred to that paper for a full
discussion. Condition 1 of an $(a,b)$-model formalizes the idea of a
hierarchical model in which there are $a+b$ parameters, $b$ of which
are effectively hyper-parameters and are fixed by the prior and the
remaining $a$ of which are model parameters.  Conditions 2 and 3
ensure realizability and an appropriate level of smoothness. 

The following definition is needed to state the main theorem.
\begin{defn}
Let $(X,\Sigma,P)$ be a measure space and $f,g\colon N\times X\to \R$
($N$ is the positive integeres) be two real-valued functions on $N
\times X$ such that for all $n\in N$, $f(n,\cdot)$ and
$g(n,\cdot)$ are measurable functions on $X$. Set $X_n := \{x\colon
f(n,x) = g(n,x)\}$ for each $n\in N$.  We say
\begin{equation}
f(n,x) \thickapprox_{(X,P)} g(n,x)
\end{equation}
if $\lim_{n\rightarrow\infty} P(X_n) = 1$. This will be abbreviated to
$f(n,x) \thickapprox g(n,x)$ when $X$ and $P$ are clear from the
context.
\end{defn}

For the following theorem, fix $n$ and take all limiting behaviour
with respect to $m$.
\begin{thm}[\cite{97b},Theorem 6]
\label{abthm}
In an $(a,b)$-model, the learner's cumulative risk \eqref{num}
satisfies
\begin{equation}
\label{burp}
\Cbar_{n,m,\pi^*} \thickapprox_{\Pi,P_\Pi} \frac{\log m}{2}\(a +
\frac{b}{n}\) + o\(\log m\).
\end{equation}
If the true prior $\pi^*$ is known then
\begin{equation}
\label{burp1}
\Cbar_{n,m,\pi^*} = \frac{\log m}{2}\(a\) + o\(\log m\).
\end{equation}
\end{thm}
Note that equation \eqref{burp1} is an equality. In \cite{97b} it was
stated as a $\doteq$ relation but in fact the stronger expression
holds.  

Comparing \eqref{burp} and \eqref{burp1}, we see that as the 
number of tasks $n$ increases, the effect of lack of knowledge of the
true prior can be made arbitrarily small. Also, learning multiple
tasks is most advantageous when $b \gg a$, \ie when the true model is
quite small but our uncertainty concerning the true model is
large. This is a similar conclusion to the one reached in section
\ref{featsec} (see note 3 after theorem \ref{ntaskthm}.

\subsection{Learning features within the Bayes model}
\label{ex1}
Consider the feature learning model of section \ref{featsec} (recall 
figure \ref{nnet}). In this case, $\Theta$ is the set of all
possible neural networks implementable by fixing the feature weights
and the weights of a single output node. As there are $k$ output
weights and $W$ feature weights, $\Theta = \R^{k+W}$. The
realizability assumption means 
there exists a fixed set of features such that all tasks in the
environment can be implemented by composing a squashed linear map with
the feature set. Thus, the true prior distribution $p(\theta|\pi^*)$
is a delta-function over the correct feature weight setting, combined
with an appropriate distribution over the $k$ output weights (one that
generates the tasks in the environment with the correct
frequency). Assuming the output-weight distribution is continuous, the
true prior is of the form \eqref{exam}, with $a=k$ and $b=W$. The set
of all such priors is simply the set of all distributions that are a
delta function over some feature weight setting, combined with a
continuous distribution over the output weights.  To complete the
model, suppose that for each $\theta\in\R^{k+W}$, $p(z|\theta)$ is of
the form
\begin{eqnarray*}
p(y=1,x|\theta) &=& p(x)f_\theta(x) \\
p(y=0,x|\theta) &=& p(x)(1-f_\theta(x))
\end{eqnarray*}
where $f_\theta(x)$ is the output of the network with weights $\theta$
and input $x$, and $p(x)$ is a continuous density on some compact
subset of $\R^d$. Assume the sigmoid on the hidden nodes is $\sigma(x) =
\tanh(x)$, and on the output node is $\sigma(x) = (1 + \tanh(x))/2$.

With these assumptions, the neural network feature learning model is
a $(k,W)$-model in the sense of definition
\ref{abdef}. Unfortunately, the neural network feature model does not
satisfy the technical conditions mentioned after definition
\ref{abdef}, so a straightforward application of theorem \ref{abthm}
is not possible. However, the difficulties are not insurmountable
(see \cite{97b} for the details) and so we obtain the following theorem:
\begin{thm}
\label{LDRthm}
For the neural-network feature learning model as above, the cumulative
risk \eqref{num} satisfies
\begin{equation}
\label{burp3}
\Cbar_{n,m,\pi^*} \thickapprox \frac{\log m}{2}\(k +
\frac{W}{n}\) + o\(\log m\).
\end{equation}
If the true prior is known (\ie the true feature weights are known),
then 
\begin{equation}
\label{burp4}
\Cbar_{n,m,\pi^*} = \frac{\log m}{2}\(k\) + o\(\log m\).
\end{equation}
\end{thm}
Note the reappearance of the factor $k + W/n$ (recall theorem
\ref{ntaskthm} and the notes afterwards). Comparing equations
\eqref{burp3} and \eqref{burp4} we see again that the effect of ignorance 
of the true prior (in this case ignorance of the right features) can
be made arbitrarily small by learning sufficiently many tasks. The
improvement is greatest when the number of features ($k$) is small,
but our uncertainty as to what the right feature set should be is
large. 

\section{Conclusion}
Two mathematical models of bias learning (or learning to learn) have
been discussed: one based on Empirical process theory and the other a
Hierarchical Bayesian model. Both models show that if a learning
machine is embedded within an {\em environment} of related learning
tasks, then it can learn its own {\em bias} for the environment by
learning sufficiently many tasks. Bounds were given on the number of
tasks and number of examples of each task needed to ensure good
generalisation from a bias learner. Good generalisation in this case
means that with high probability the learner's choice of hypothesis
space will contain good solutions to novel tasks drawn from the same
environment. The theory was specialised to the case of feature
learning with neural networks. 

There are many pattern classification problems that can be viewed as
consisting of large number of related tasks and that seem to possess
small feature sets. Speech recognition, character recognition, face
recognition and fingerprint recognition all fit this bill. Feature
learning in these domains should be particularly successful.
   
     
\bibliographystyle{plain}
\bibliography{bib}
     
\end{article}
\end{document}